# Self-supervised contrastive learning of echocardiogram videos enables label-efficient cardiac disease diagnosis

Gregory Holste[1], Evangelos K. Oikonomou[2], Bobak J. Mortazavi[3], Zhangyang Wang[1], Rohan Khera[2]

[1] Department of Electrical and Computer Engineering, The University of Texas at Austin, Austin, TX, USA
[2] Section of Cardiovascular Medicine, Department of Internal Medicine, Yale School of Medicine, New Haven, CT, USA
[3] Department of Computer Science & Engineering, Texas A&M University, College Station, TX, USA

**Address for correspondence:**
Rohan Khera, MD, MS
195 Church St, 6th Floor, New Haven, CT 06510
203-764-5885; rohan.khera@yale.edu; @rohan_khera

## ABSTRACT

Advances in self-supervised learning (SSL) have shown that self-supervised pretraining on medical imaging data can provide a strong initialization for downstream supervised classification and segmentation. Given the difficulty of obtaining expert labels for medical image recognition tasks, such an "in-domain" SSL initialization is often desirable due to its improved label efficiency over standard transfer learning. However, most efforts toward SSL of medical imaging data are not adapted to *video-based* medical imaging modalities. With this progress in mind, we developed a self-supervised contrastive learning approach, **EchoCLR**, catered to echocardiogram videos with the goal of learning strong representations for efficient fine-tuning on downstream cardiac disease diagnosis. EchoCLR leverages (i) *distinct* videos of the same patient as positive pairs for contrastive learning and (ii) a frame re-ordering pretext task to enforce temporal coherence. When fine-tuned on small portions of labeled data (as few as 51 exams), EchoCLR pretraining significantly improved classification performance for left ventricular hypertrophy (LVH) and aortic stenosis (AS) over other transfer learning and SSL approaches across internal and external test sets. For example, when fine-tuning on 10% of available training data (519 studies), an EchoCLR-pretrained model achieved 0.72 AUROC (95% CI: [0.69, 0.75]) on LVH classification, compared to 0.61 AUROC (95% CI: [0.57, 0.64]) with a standard transfer learning approach. Similarly, using 1% of available training data (53 studies), EchoCLR pretraining achieved 0.82 AUROC (95% CI: [0.79, 0.84]) on severe AS classification, compared to 0.61 AUROC (95% CI: [0.58, 0.65]) with transfer learning. EchoCLR is unique in its ability to learn representations of medical videos and demonstrates that SSL can enable label-efficient disease classification from small, labeled datasets.

## INTRODUCTION

Transthoracic echocardiography (TTE) represents ultrasound videos that assess cardiac structure and function. For nearly all structural and functional cardiac diseases, TTE is the gold-standard for diagnosis due to its high temporal resolution and ability to capture multiple viewpoints of a patient's beating heart flexibly.[1,2] Recently, deep learning techniques have been developed and adapted to detect various cardiac diseases from echocardiography.[3–8] While important steps forward, nearly all of these works either (a) examine individual still frames, ignoring the temporal aspect of echocardiography, or (b) rely on standard deep supervised learning for videos, which requires large labeled datasets for adequate training. The latter is particularly burdensome for clinical diagnosis tasks, where acquiring expert labels for tens of thousands (or more) medical videos can be prohibitively time-consuming and expensive.[9] Therefore, we sought to develop a "label-efficient" self-supervised learning (SSL) approach to learn representations of echocardiogram videos as powerful initializations for downstream fine-tuning on small, labeled datasets. We validate our proposed approach on the classification of left ventricular hypertrophy (LVH), an abnormal thickening of the left ventricle walls that is measurable from a single TTE video, and aortic stenosis (AS), a disorder causing narrowing of the aortic valve typically diagnosed with complex multi-view and Doppler echocardiography.

SSL has proven very effective for label-efficient fine-tuning in natural image classification,[10,11] video classification,[12,13] and now even medical image classification and segmentation tasks.[14–17] However, most successful medical applications of SSL operate on 2D data such as histopathological images and radiographs.[16,18,19] Some recent studies have developed SSL methods for 3D medical image data, though this has typically been applied to computed tomography (CT) and magnetic resonance imaging (MRI), where this third dimension is *spatial*, not temporal.[15,20] That is, medical videos should require special treatment to handle the structure of temporal content – for TTE, the spatial translocation of cardiac structures over time. Also, virtually all applications of SSL to 3D medical imaging are in high-resolution modalities with standardized acquisition, making them amenable to the data augmentation required to power many SSL approaches. In contrast, ultrasound is a noisy modality, acquired manually by a sonographer with frequent motion during acquisition, producing low-resolution images that are sensitive to transformations that require pixel interpolation. Moreover, echocardiography is highly dependent on the skill of the sonographer, with frequent off-axis images containing distortions and artifacts.[21]

There are two main challenges to applying SSL – specifically, contrastive learning[10,11] – to echocardiograms: (i) ultrasound is brittle to heavy data augmentation due to relatively low spatial resolution and contrast, and (ii) echocardiograms contain rich temporal content that most contrastive learning approaches ignore. To tackle the first issue, we employ *"multi-instance" contrastive learning*. In the setting of TTE, a patient often will have multiple videos captured from a single view type during routine acquisition; we can then leverage these different instances of the same canonical video type for a given patient as "positive pairs," of which the model will learn to produce similar representations. Assuming that different videos of the same patient capture the same underlying characteristics, this forms challenging positive pairs for contrastive representation learning and removes the need for aggressive augmentation to artificially generate two distinct "views" of a patient. To address the second issue of incorporating the temporal nature of echocardiography, we utilize a pretext task of frame re-ordering. For this, we randomly shuffle the frames of an input echo and train a classifier to predict the original order of frames. This imbues the model with a sense of temporal coherence that we observe to aid downstream fine-tuning for severe AS and LVH classification and generate more interpretable visual explanations of these predictions.

Through validation on internal and external testing data, we demonstrate that the proposed method, **EchoCLR**, significantly outperforms existing transfer learning and SSL methods for both LVH and severe AS classification when trained on a small number of labeled TTE studies (as few as 51). While existing work applies contrastive learning to echocardiography,[22–24] these efforts only operate on individual still frames and do not perform downstream disease classification. To our knowledge, this represents the first effort toward self-supervised contrastive learning of echocardiogram videos for label-efficient cardiac disease diagnosis.

## RESULTS

### Study cohort description

This study included patients who underwent transthoracic echocardiograms (TTE) between the years 2016 and 2021 at the Yale New Haven Hospital, with LVH and severe AS labels determined from cardiologist interpretation. A sample of 10,000 studies from 2016-2020 was extracted for model development and internal testing, while a sample of 2,500 studies from 2021 was extracted to serve as a temporally distinct external test set. To obtain videos solely from the parasternal long axis (PLAX) view – the first and most common view obtained in TTE – a pretrained automatic view classifier was employed.[25] Following view classification, the resulting PLAX videos underwent deidentification by using an image processing pipeline to mask pixels outside of the central image content. The resulting 2016-2020 cohort consisted of 23,448 videos from 7,082 TTE studies, while the temporally distinct 2021 cohort consisted of 6,530 videos from 2,040 studies. While AS labels were present for all studies, LVH labels could only be determined for a subset of 22,952 videos from 6,931 studies in the 2016-2020 cohort and 6,379 videos from 1,995 studies in the 2021 cohort. Demographic information can be found in **Table 1** and full data curation and preprocessing details can be found in the **Methods**.

### Label-efficient LVH classification

To evaluate label efficiency, we compared three different initialization strategies (random weights, Kinetics-400 weights, and EchoCLR-pretrained weights) and fine-tuned each model on increasing amounts of training data, ranging from 1% to 100% of available data. When using all available training data (5,194 studies), EchoCLR pretraining and Kinetics-400 pretraining, representing a standard video-based transfer learning approach, classified LVH comparably. With all training data, an EchoCLR-pretrained model reached 0.795 AUROC (95% CI: [0.770, 0.819]) on the internal test set and 0.804 AUROC (95% CI: [0.783, 0.824]) on the external test set, compared to 0.807 AUROC (95% CI: [0.783, 0.830], P=0.818) in internal testing and 0.806 AUROC (95% CI: [0.786, 0.827], P=0.599) in external testing with Kinetics-400 pretraining.

However, when using less than 25% of training data for fine-tuning (<1,000 studies), EchoCLR pretraining consistently outperformed other initialization methods on downstream LVH classification, as measured by AUROC in internal and external test sets (**Fig. 2A, 2B**). For example, when using only 10% of training data for fine-tuning (519 studies), an EchoCLR-pretrained model reached 0.723 AUROC (95% CI: [0.693, 0.751]), significantly outperforming the Kinetics-400-pretrained model (0.605 AUROC, 95% CI: [0.574, 0.636], P<0.001) and randomly initialized model (0.488 AUROC, 95% CI: [0.456, 0.521], P<0.001); in the external test set, the EchoCLR-pretrained model reached 0.701 AUROC (95% CI: [0.676, 0.725]), again outperforming the Kinetics-400-pretrained model (0.605 AUROC, 95% CI: [0.578, 0.632], P<0.001) and randomly initialized model (0.493 AUROC, 95% CI: [0.465, 0.520], P<0.001). Overall, EchoCLR pretraining significantly improved upon Kinetics-400 transfer learning on 1%, 5%, 10%, and 50% training ratios in the internal test set and all training ratios except 100% in the external test set. These same trends were also reflected in performance by AUPR (**Fig. S1A, S1B**), and full results for LVH classification can be found in **Table S2**.

**Label-efficient severe AS classification**

Aligned with the patterns observed in LVH classification, EchoCLR and Kinetics-400 initializations performed similarly for downstream severe AS classification. Leveraging all available fine-tuning data (5,311 studies), EchoCLR pretraining achieved 0.934 AUROC (95% CI: [0.920, 0.947]) on the internal test set and 0.947 AUROC (95% CI: [0.884, 0.988]) on the external test set, while Kinetics-400 pretraining reached 0.938 AUROC (95% CI: [0.925, 0.951], P=0.774) in internal testing and 0.954 (95% CI: [0.921, 0.983], P=0.615) in external testing.

Once again, however, EchoCLR pretraining improved severe AS classification when leveraging relatively small portions of labeled data (<1,000 labeled TTE studies) (**Fig. 1C, 1D**). For example, when fine-tuned on only 1% of training data (53 studies), our EchoCLR-pretrained model reached 0.818 AUROC (95% CI: [0.793, 0.840]) in the internal test set, significantly outperforming a Kinetics-400-pretrained model (0.612 AUROC, 95% CI: [0.577, 0.647], P<0.001) and randomly initialized model (0.511 AUROC, 95% CI: [0.477, 0.545], P<0.001). Further, in external testing, the EchoCLR-pretrained model reached 0.874 AUROC (95% CI: [0.820, 0.922]), outperforming the Kinetics-400-pretrained model (0.645 AUROC, 95% CI: [0.525, 0.761], P<0.001) and randomly initialized model (0.437 AUROC, 95% CI: [0.338, 0.535], P<0.001). Overall, EchoCLR pretraining significantly outperformed the standard transfer learning approach of Kinetics-400 pretraining on 1%, 5%, and 10% training ratios in both the internal and external test sets. In other words, once fine-tuning on at least 25% of training data (> 1,000 studies), all initialization methods were comparable in terms of downstream severe AS classification performance. Consistent trends were also observed with respect to AUPR (**Fig. S1C, S1D**); see **Table S3** for full results.

**Ablation study on EchoCLR**

To demonstrate the additive impact of each component of EchoCLR, we compared the downstream classification performance of EchoCLR pretraining, EchoCLR without frame reordering ("multi-instance" MI-SimCLR), and EchoCLR without frame reordering and without multi-instance sampling (standard SimCLR). EchoCLR pretraining considerably improved both LVH and severe AS classification over other self-supervised pretraining approaches when fine-tuning on the vast majority of training set ratios. Additionally, the proposed multi-instance echocardiography sampling and frame re-ordering provided *complementary* improvements to downstream classification performance (**Fig. 3**).

To illustrate this, when using just 1% of data for severe AS fine-tuning (53 studies), the EchoCLR-pretrained model reached 0.818 AUROC (95% CI: [0.793, 0.840]), significantly outperforming an MI-SimCLR-pretrained model (0.718 AUROC, 95% CI: [0.685, 0.751],

P<0.001) and SimCLR-pretrained model (0.569 AUROC, 95% CI: [0.534, 0.604], P<0.001); in the external test set, the EchoCLR-pretrained model achieved 0.874 AUROC (95% CI: [0.820, 0.922]), outperforming both the MI-SimCLR-pretrained model (0.702 AUROC, 95% CI: [0.610, 0.795], P<0.001) and SimCLR-pretrained model (0.528 AUROC, 95% CI: [0.427, 0.628], P<0.001). Similar trends were observed for severe AS classification performance measured by AUPR (**Table S2**) as well as LVH classification (**Fig. 3**), as measured by both AUROC and AUPR in internal and external test sets (**Table S3**).

**Improved interpretability with EchoCLR**

In addition to evaluating quantitative disease classification performance, we perform a comparative interpretability analysis of different pretraining methods. To ensure that clinically relevant regions of the patient's imaging contribute to the model's predictions, saliency maps were generated for the four most confident severe AS predictions for the Kinetics-400-pretrained and EchoCLR-pretrained model (utilizing all available training data) with GradCAM.[26] The EchoCLR-pretrained model more closely attended to the aortic valve and annulus than the Kinetics-400-pretrained model, whose heatmaps were more diffuse, capturing regions that should not necessarily be relevant to severe AS diagnosis. Refer to the **Methods** for full details on saliency map generation and visualization.

# DISCUSSION

We developed EchoCLR, the first *spatiotemporal* self-supervised contrastive learning method specifically catered to echocardiography, a key video-based medical imaging modality. Through extensive internal and external validation, we showed that EchoCLR pretraining improved downstream classification for a diverse set of complex classification tasks by 5-40% when fine-tuned on small amounts of labeled data (<1,000 TTE studies) compared to other deep learning initialization approaches. In the "low-data regime," EchoCLR pretraining consistently outperformed randomly initializing weights and the standard video-based transfer learning with Kinetics-400-pretrained weights on both LVH and severe AS classification. These results demonstrate the label efficiency of "in-domain" self-supervised pretraining on echocardiograms with EchoCLR, enabling accurate disease classification from small labeled echocardiographic datasets.

Further, our ablation study on EchoCLR demonstrates that the label efficiency does not merely come from pretraining on echocardiograms, but rather the modality-informed modifications proposed in EchoCLR: multi-instance echocardiography sampling and frame re-ordering. It was hypothesized that (i) using *distinct* videos of the same patient as positive pairs for contrastive learning would remove the need for heavy augmentation and enable more effective representation learning, and (ii) using a frame re-ordering pretext task would enforce temporal coherence that existing methods like SimCLR could not. These hypotheses were supported by the observation that MI-SimCLR (SimCLR with multi-instance sampling) improved upon SimCLR (no multi-instance sampling), and EchoCLR (MI-SimCLR with frame reordering) improved upon MI-SimCLR across independent downstream disease diagnosis tasks over nearly all fine-tuning ratios in both internal and external testing.

Despite the benefits of self-supervised contrastive pretraining when fine-tuning on small amounts of labeled data, our experiments reveal a "saturation point" of approximately 25% of training data (>1,000 TTE studies), after which all initialization methods were usually comparable (no significant difference in downstream classification performance). Other works on self-supervised learning observe this behavior,[27,28] which may be explained by the fact that while in-domain SSL may enable more rapid adaptation to a downstream task, a sufficient number of labeled examples may provide enough signal for any reasonable initialization to reach a similar minimum in the loss landscape. However, several studies on 2D medical image-based applications of SSL have found that SSL pretraining can even outperform fully supervised approaches in the presence of large amounts of labeled data.[14,29,30]

Future work may explore ways to combine contrastive pretraining with other types of SSL such as generative modeling[31,32] and masked autoencoding.[29,30,33] While this study was limited to single-view echocardiography – to isolate the effect of EchoCLR on two key diagnostic tasks from one of the most widely acquired echocardiographic views – developing an SSL method for multi-view echocardiography would allow for even richer representation learning. Additionally, these SSL approaches can further be applied to multimodal data, learning holistic joint representations across modalities, such as between paired echocardiograms and other diagnostic cardiovascular assessments. Methods like EchoCLR have the potential to accelerate deep learning for low-prevalence disease detection, as well as any clinical setting where it is difficult to acquire large-scale, expert-labeled medical imaging datasets.

## METHODS

### Data curation and deidentification
An initial pull of 12,500 TTE studies conducted at Yale New Haven Hospital from 2016-2021 was extracted for this study. To ensure sufficient positive examples for model development and reliable evaluation, the set of 10,000 studies from 2016-2020 was enriched for AS prevalence by oversampling mild, mild-moderate, moderate, and moderate-severe AS by a factor of 5 and oversampling severe AS by a factor of 50. The remaining 2,500 studies from 2021, representing a temporally distinct external test set, were *not* enriched for AS. Of the 10,865 studies that were properly extracted from the database, 9,710 studies contained valid pixel data in the DICOM files. The resulting 447,653 videos were then deidentified by masking the periphery of pixels in each frame to remove protected health information, then converted to Audio Video Interleave (AVI) format. The view classification and video preprocessing details below follow those described in our previous work,[36] though the cohort of studies used in this work differs.

### View Classification
This study leveraged single-view echocardiography from the PLAX view by employing a pre-trained TTE view classifier developed by Zhang *et al.*[25] For all 447,653 videos, ten deidentified frames were randomly selected, downsampled to 224 x 224 resolution, and fed into the view classifier. The 10 frame-level predicted view probabilities were then averaged into a single video-level view prediction. While the pretrained view classifier was capable of discriminating variants of the canonical PLAX view (e.g., "PLAX – zoom of left atrium," "PLAX – remote," etc.), we only retrained videos most confidently classified as "PLAX."

**Preprocessing**
After view classification, the 30,136 videos from 9,173 studies were prepared for deep learning model development. Given differences in AS severity measures across different domains, we excluded echocardiograms with low-flow, low-gradient, and paradoxical AS, leaving 29,978 PLAX videos from 9,122 studies. All videos underwent a more thorough cleaning and deidentification process that involved binarizing each video frame with a fixed threshold of 200, then masking out all pixels outside the convex hull of the largest contour to remove all information outside the central image content. Finally, each video clip was spatially downsampled to 112 x 112 studies and saved to AVI format for fast loading during model training. All videos from studies conducted from 2016-2020 were then randomly split at the study level into training (75%), validation (10%), and internal test (15%) sets. See **Table S1** for detailed demographic details of the study cohort.

**Echocardiogram labeling**
Echocardiographic measurements and reported diagnoses are reported in accordance with the recommendations of the American Society of Echocardiography (ASE).[1,34,35] Specifically, the presence of AS severity was adjudicated based on the original echocardiographic report and reflected the final severity grade assigned by the interpreting physician. Since severe AS detection was formulated as a binary classification task, all AS designations other than "severe AS" were binned into the "not severe AS" category. LVH was defined based on sex-specific thresholds for the left ventricular mass index, namely >95 g/m$^2$ in women and >115 g/m$^2$ in men.[34]

**Self-supervised pretraining**
To learn transferable representations of PLAX echocardiogram videos for downstream cardiac disease classification, we performed self-supervised pretraining on all training set videos. This pretraining step enables the model to learn representations of echocardiograms that are robust to variations in video acquisition and more rapidly adapt to the target fine-tuning task (based on fewer training examples) than other initialization approaches. To this end, we designed an SSL algorithm specifically catered to echocardiogram videos, **EchoCLR**, which consists of a novel combination of (i) a multi-instance contrastive learning task and (ii) a frame re-ordering pretext task.

We adopted the SimCLR framework[10] for contrastive learning, which produces two "views" of an image by sending two copies of the input through a pipeline of random image augmentations, producing view $\tilde{x}_i$ and $\tilde{x}_j$. An encoder $f()$ is then used to learn representations of each view, $h_i = f(\tilde{x}_i)$ and $h_j = f(\tilde{x}_j)$, which are then projected to a lower dimensionality with a projector $g()$. The resulting learned embeddings of each view, $z_i = g(f(\tilde{x}_i))$ and $z_j = g(f(\tilde{x}_j))$ are then "contrasted" via the temperature-normalized cross-entropy (NT-Xent) loss, which encourages the model to learn *similar* representations of views from the same original image (so-called "positive pairs") and *dissimilar* representations of views from all other images ("negative pairs") in a given minibatch. This loss is computed via

$$\ell_{i,j} = -log\frac{exp(sim(z_i, z_j)/\tau)}{\sum_{k=1}^{2N} \mathbb{1}_{[i \neq k]} exp(sim(z_i, z_k)/\tau)},$$

where $sim(\cdot)$ is the cosine similarity and $\tau$ is a "temperature" hyperparameter; the final loss is then simply the average $\ell_{i,j}$ for all positive pairs $(i, j)$ and $(j, i)$ in the minibatch.

While SimCLR has proven successful for 2D natural images as well as medical modalities such as radiography and histopathology,[16,18,19] there are several barriers to its successful adaptation to echocardiography. First, SimCLR requires heavy image augmentation for effective representation learning, which would destroy valuable signal encoded in the noisy, low-contrast ultrasound images produced by echocardiography. Second, SimCLR was designed for 2D images, completely ignoring the temporal dimension of echocardiography. To address the first issue, we utilized "multi-instance" contrastive learning – borrowing language and key insights from Azizi *et al.*[37] – whereby we form positive pairs between *different* videos from the *same* patient. This critically removes the need to synthetically create two different "views" of a patient by heavily augmenting their echocardiogram, instead leveraging the fact that almost all routine TTE studies contain multiple distinct PLAX videos of a patient. Throughout the main text, we refer to this method as MI-SimCLR (multi-instance SimCLR). To address the second issue, we additionally included a frame re-ordering "pretext" task on top of MI-SimCLR, where we randomly permuted the frames of each input echocardiogram, then trained the model to predict the shuffled order of frames to enforce temporal coherence within a cardiac cycle. Similar to the approach of Jiao *et al.*,[38] this frame re-ordering task is treated as a classification problem and was implemented with a fully-connected layer that minimizes the cross-entropy between the known and predicted original frame order; specifically, if an input echo clip has $K$ frames, then the $K!$ possible permutations of frames served as the targets for classification. The final loss function of EchoCLR is simply the sum of the contrastive NT-Xent objective and the pretext frame re-ordering cross-entropy objective.

Self-supervised pretraining was performed on randomly sampled video clips of $K = 4$ consecutive frames from all training set echocardiogram videos. The encoder $f()$ was a randomly initialized 3D-ResNet18,[39] and the projector $g()$ projected each 512-dimensional learned representation down to a 128-dimensional representation with a hidden layer of 256 units followed by a ReLU activation, followed by an output layer with 128 units. Each model used the Adam optimizer,[40] a learning rate of 0.1, a batch size of 392 (196 per GPU), and an NT-Xent temperature hyperparameter of 0.5. The following augmentations were applied to each frame in a temporally consistent manner (same transformations for each frame of a given video clip): random zero padding by up to 8 pixels in each spatial dimension, a random horizontal flip with probability 0.5, and a random rotation within -10 and 10 degrees with probability 0.5; for EchoCLR pretraining, each video clip was additionally permuted along the temporal dimension. After augmentation, each video clip was normalized so that the maximum pixel intensity was mapped to 1 and the minimum intensity to 0. For methods that used multi-instance echocardiography sampling (MI-SimCLR and EchoCLR), the model was trained for 300 epochs on all unique pairs of *different* PLAX videos acquired in the same study. Since SimCLR does not use multi-instance sampling, this method was trained for 520 epochs to match the number of optimization steps (or "examples seen") during MI-SimCLR and EchoCLR pretraining for a fair comparison. Each model was trained on two NVIDIA RTX 3090 GPUs.

**Supervised fine-tuning**

The same 3D-ResNet18 architecture was used to classify LVH and severe AS. Three different methods were used to initialize the parameters of this network: an SSL initialization (using one of SimCLR, MI-SimCLR, or EchoCLR), a Kinetics-400 initialization, and a random initialization. The SSL initializations directly used the learned weights of the encoder from the one of the three SSL pretraining approaches described in detail above. The Kinetics-400 initialization represents a "standard" transfer learning approach for spatiotemporal data, using the weights from a 3D-ResNet18 trained in a supervised fashion on the Kinetics-400 dataset, a large corpus of over 300,000 natural videos for human action classification; these weights are readily available through the *torchvision* API (https://pytorch.org/vision/stable/index.html) provided by PyTorch. The random initialization is the default when initializing a 3D-ResNet18 with PyTorch.[41] To understand how each initialization compared with respect to label efficiency, a training ratio (or "data titration") experiment was performed: for each outcome and initialization method, a 3D-ResNet18 was initialized with the given method and fine-tuned on 1%, 5%, 10%, 25%, 50%, and 100% of all available training set TTE studies for the given outcome. For LVH classification, this involved using 51 (1%), 259 (5%), 519 (10%), 1,298 (25%), 2,597 (50%), and 5,194 (100%) studies for fine-tuning; for severe AS, this involved 53 (1%), 265 (5%), 531 (10%), 1,327 (25%), 2,655 (50%), and 5,311 (100%) studies.

All fine-tuned models were trained on randomly sampled clips of 16 consecutive frames from training set echocardiograms, optionally padding with empty frames if needed. The same augmentations were used as in self-supervised pretraining, and all video clips were min-max normalized as described above; when fine-tuning from a Kinetics-400 initialization, video clips were further standardized using the channel-wise means and standard deviations from the Kinetics-400 training dataset, a standard preprocessing step when performing transfer learning. All models were trained for a maximum of 30 epochs with early stopping – specifically, if the validation loss did not improve for 5 consecutive epochs, training was terminated and the weights from the epoch with minimum validation loss were used for final evaluation. All fine-tuned models were trained on a single NVIDIA RTX 3090 GPU with the Adam optimizer and a batch size of 88 to maximize GPU utilization. When fine-tuning on more than 10% of available training data, the Kinetics-pretrained and randomly initialized models used a learning rate of $1 \times 10^{-4}$, while the SSL-pretrained models used a learning rate of 0.1. When fine-tuning on 10% or less of available data, the Kinetics-pretrained and randomly initialized models used a learning rate of $1 \times 10^{-4}$, $5 \times 10^{-5}$, or $1 \times 10^{-5}$, while the SSL-pretrained models used a learning rate of 0.1, 0.05, or 0.001; this was done due to the stochasticity introduced by training on a very small random subset of data, and the learning rate with minimum validation loss at the early stopping-determined checkpoint was ultimately selected.

**Model explainability**

Saliency maps were generated by leveraging the Grad-CAM method[42] for obtaining visual explanations from deep neural networks. Specifically, heatmaps were generated by applying Grad-CAM to a clip of the first 32 frames of an echo, using the last convolution block of the 3D-ResNet18 to generate a 7 x 7 x 4 (height x width x time) heatmap displaying roughly where the model is attending to over the spatial and temporal dimensions. The Grad-CAM output was interpolated to the original input dimension of 112 x 112 x 32 with the scipy "zoom" function; this process produces a frame-by-frame "visual explanation" of *where* (in space and time) the model is focusing frame by frame in order to make its prediction. However, to generate a single 2D heatmap for a given echo clip, the pixelwise maximum along the temporal dimension was taken to capture the most salient regions for severe AS presence across all timepoints.

**Evaluation and statistical analysis**

Since severe AS and LVH labels describe each TTE study, but the model is trained on multiple echocardiogram video clips from the same study, video-level predictions of disease presence are averaged at inference time to obtain a single study-level prediction. All fine-tuned severe AS and LVH classification models were evaluated by AUROC and area under the precision-recall curve (AUPR), which can be useful when evaluating a classifier's performance in the presence of imbalanced data. Full detailed results for each outcome, training set ratio, and initialization method can be found in the **Supplemental Materials**.

All 95% confidence intervals for model evaluation metrics were computed by bootstrapping. At the study level, 10,000 bootstrap samples (samples with replacement of the same size as the evaluation set) were drawn, metrics were computed on this bootstrapped set, and nonparametric confidence intervals were generated with the percentile method.[43] All p-values were computed with a Python implementation of the "bootstrap" method described in the documentation for the *roc.test* function (https://cran.r-project.org/web/packages/pROC/pROC.pdf) in the pROC library.[44] Significance tests were one-sided – the null hypothesis being that the EchoCLR-pretrained model's AUROC exceeded the Kinetics-400-pretrained model's AUROC – with significance level 0.05.

**DATA AVAILABILITY**

The data used in this study are not available for public sharing given the restrictions in our institutional review board approval. Deidentified test data may be made available to researchers under a data use agreement after publication in a peer-reviewed journal.

**CODE AVAILABILITY**

The code repository for this work can be found at https://github.com/CarDS-Yale/EchoCLR.

**Table 1 | Description of study cohort**

| | | Overall | Training | Validation | Internal Testing | External Testing |
|---|---|---|---|---|---|---|
| **TTE Studies, n** | | 9,122 | 5,311 | 708 | 1,063 | 2,040 |
| **Age (years), mean (SD)** | | 69.1 (16.0) | 70.2 (15.8) | 70.1 (15.6) | 69.8 (15.8) | 65.7 (16.4) |
| **Gender, n (%)** | Female | 4,467 (49.0) | 2,600 (49.0) | 349 (49.3) | 521 (49.0) | 997 (48.9) |
| | Male | 4,655 (51.0) | 2,711 (51.0) | 359 (50.7) | 542 (51.0) | 1,043 (51.1) |
| **Race, n (%)** | Asian | 120 (1.3) | 60 (1.1) | 6 (0.8) | 14 (1.3) | 40 (2.0) |
| | African American | 836 (9.2) | 468 (8.8) | 68 (9.6) | 96 (9.0) | 204 (10.0) |
| | Other | 498 (5.5) | 272 (5.1) | 35 (4.9) | 56 (5.3) | 135 (6.6) |
| | Unknown | 929 (10.2) | 487 (9.2) | 80 (11.3) | 114 (10.7) | 248 (12.2) |
| | White/Caucasian | 6,739 (73.9) | 4,024 (75.8) | 519 (73.3) | 783 (73.7) | 1,413 (69.3) |
| **BMI (kg/m^2), mean (SD)** | | 29.5 (16.3) | 29.4 (19.6) | 30.1 (16.7) | 29.4 (8.2) | 29.4 (7.3) |
| **Severe Aortic Valve Stenosis, n (%)** | | 1,609 (17.6) | 1,183 (22.3) | 160 (22.6) | 246 (23.1) | 20 (1.0) |
| **Left Ventricular Hypertrophy, n (%)** | | 2,280 (25.5) | 1,398 (26.9) | 199 (28.6) | 302 (29.0) | 381 (19.1) |

Descriptive statistics of demographics and label prevalence for each set of the study cohort. Percentages are "valid percentages" calculated for studies with available information. BMI = body mass index; SD = standard deviation; TTE = transthoracic echocardiography.

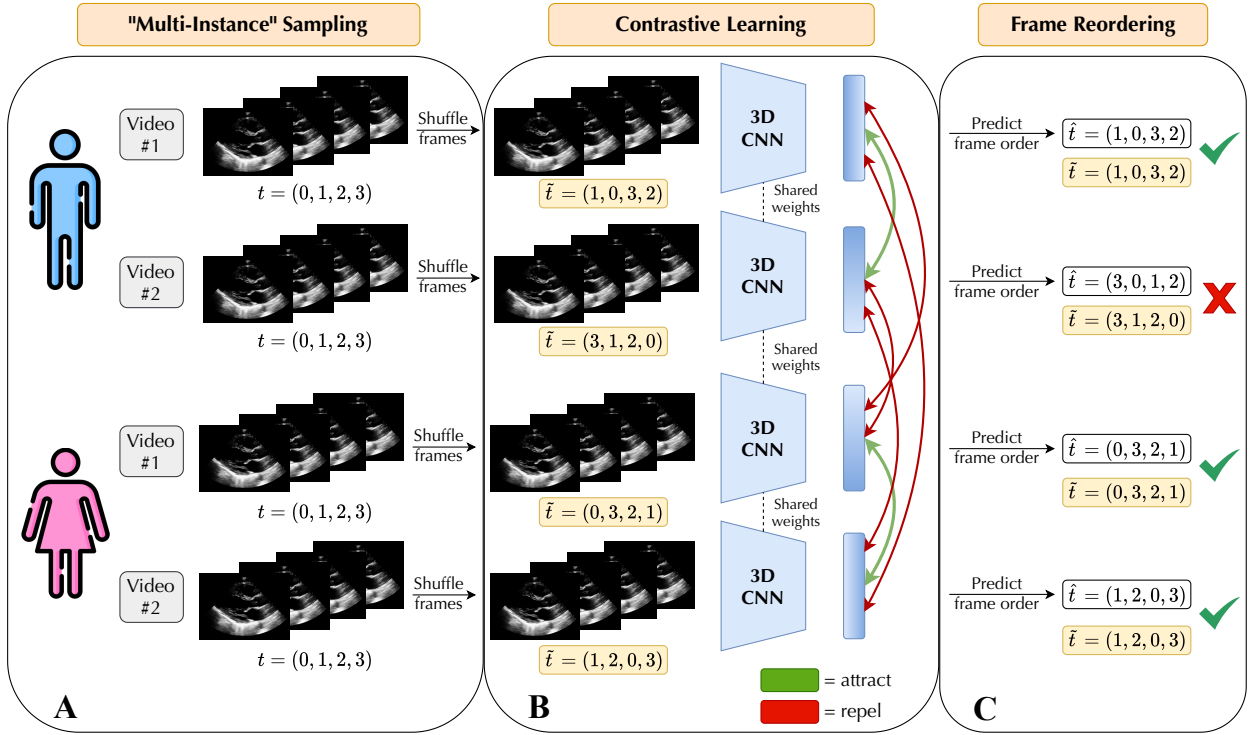

**Figure 1 | Overview of EchoCLR, a self-supervised learning approach for echocardiography.** Unlike standard contrastive learning methods, two *distinct* videos of each patient acquired during a single exam are randomly sampled and deemed positive pairs for "multi-instance" contrastive learning **(A)**. The frames of each video are then randomly shuffled along the temporal axis and fed into a 3D CNN, which learns *similar* representations of distinct videos from the same patient and *dissimilar* representations of videos from different patients **(B)**. These video-level representations are then used to directly predict the order of shuffled video frames. This frame reordering pretext task encourages temporal coherence, which we demonstrate to be beneficial for downstream echocardiogram video-based disease classification tasks **(C)**. After self-supervised pretraining, the 3D CNN backbone can then be efficiently fine-tuned for cardiac disease classification based on very few labeled echocardiograms. CNN = convolutional neural network.

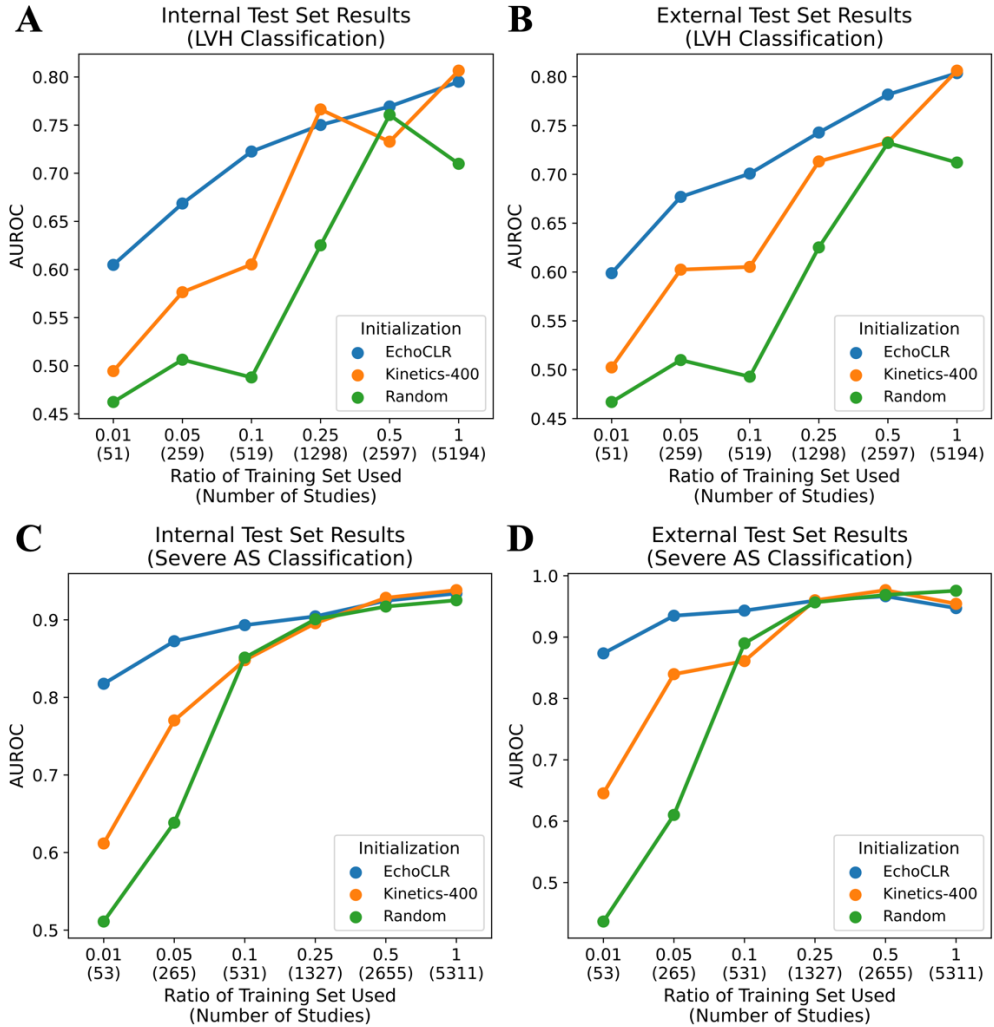

**Fig. 2 | Classification performance on different amounts of training data.** AUROC for LVH classification on the internal **(A)** and external test set **(B)** and severe AS classification on the internal **(C)** and external test set **(D)** for a randomly initialized, Kinetics-400-pretrained, and EchoCLR-pretrained model when fine-tuned on different amounts of labeled training data. AS = aortic stenosis; AUROC = area under the receiver operating characteristic curve; LVH = left ventricular hypertrophy.

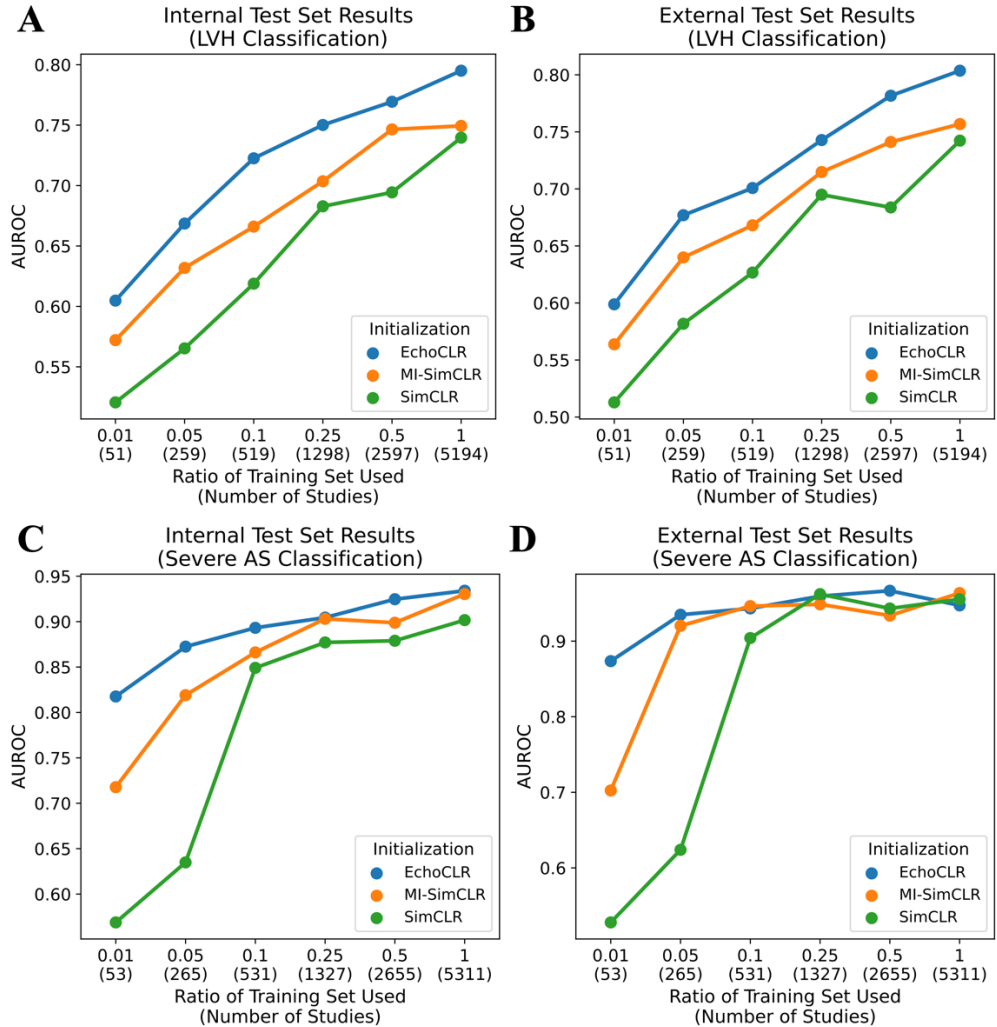

**Fig. 3 | Ablation study of EchoCLR when finetuned on different amounts of training data.** AUROC for LVH classification on the internal **(A)** and external test set **(B)** and severe AS classification on the internal **(C)** and external test set **(D)** for a SimCLR-pretrained, MI-SimCLR-pretrained, and EchoCLR-pretrained model when fine-tuned on different amounts of labeled training data. AS = aortic stenosis; AUROC = area under the receiver operating characteristic curve; LVH = left ventricular hypertrophy; MI-SimCLR = multi-instance SimCLR.

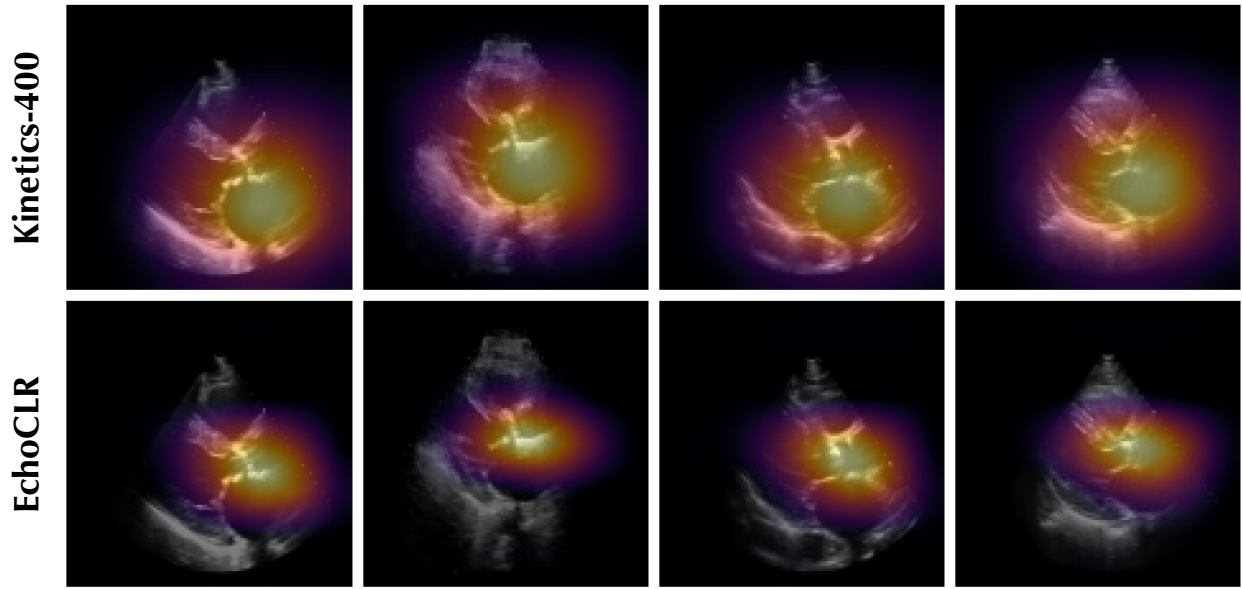

**Fig. 4 | Interpretability analysis of EchoCLR.** Saliency map visualizations for the four most confident severe AS predictions as determined by the Kinetics-400-pretrained model when fine-tuned on all training data. For each video, attention maps are presented for the Kinetics-400-pretrained model **(top row)** and the EchoCLR-pretrained model **(bottom row)**. While saliency maps were computed for entire video clips, only the first frame of each video is displayed above in each column. Saliency maps were obtained by applying Grad-CAM[26] to the first 32 frames of each video, then computing the pixelwise maximum along the temporal axis to obtain a single 2D heatmap.

**Table S1 | Detailed description of study cohort**

| | | Overall | Training | Validation | Internal Testing | External Testing |
|---|---|---|---|---|---|---|
| **TTE Studies, n** | | 9,122 | 5,311 | 708 | 1,063 | 2,040 |
| **Age (years), mean (SD)** | | 69.1 (16.0) | 70.2 (15.8) | 70.1 (15.6) | 69.8 (15.8) | 65.7 (16.4) |
| **Gender, n (%)** | **Female** | 4,467 (49.0) | 2,600 (49.0) | 349 (49.3) | 521 (49.0) | 997 (48.9) |
| | **Male** | 4,655 (51.0) | 2,711 (51.0) | 359 (50.7) | 542 (51.0) | 1,043 (51.1) |
| **Race, n (%)** | **Asian** | 120 (1.3) | 60 (1.1) | 6 (0.8) | 14 (1.3) | 40 (2.0) |
| | **African American** | 836 (9.2) | 468 (8.8) | 68 (9.6) | 96 (9.0) | 204 (10.0) |
| | **Other** | 498 (5.5) | 272 (5.1) | 35 (4.9) | 56 (5.3) | 135 (6.6) |
| | **Unknown** | 929 (10.2) | 487 (9.2) | 80 (11.3) | 114 (10.7) | 248 (12.2) |
| | **White/Caucasian** | 6,739 (73.9) | 4,024 (75.8) | 519 (73.3) | 783 (73.7) | 1,413 (69.3) |
| **Ethnicity, n (%)** | **Hispanic or Latino** | 633 (6.9) | 344 (6.5) | 51 (7.2) | 80 (7.5) | 158 (7.7) |
| | **Non-Hispanic** | 7,266 (79.7) | 4,295 (80.9) | 562 (79.4) | 848 (79.8) | 1,561 (76.5) |
| | **Unknown** | 1,223 (13.4) | 672 (12.7) | 95 (13.4) | 135 (12.7) | 321 (15.7) |
| **BMI (kg/m^2), mean (SD)** | | 29.5 (16.3) | 29.4 (19.6) | 30.1 (16.7) | 29.4 (8.2) | 29.4 (7.3) |
| **EF (%), mean (SD)** | | 58.9 (10.6) | 59.0 (10.8) | 59.2 (10.7) | 58.6 (10.8) | 59.0 (10.1) |
| **Aortic Valve Stenosis, n (%)** | **Sclerosis without Stenosis** | 906 (9.9) | 471 (8.9) | 62 (8.8) | 90 (8.5) | 283 (13.9) |
| | **Mild** | 962 (10.5) | 668 (12.6) | 79 (11.2) | 132 (12.4) | 83 (4.1) |
| | **Moderate** | 634 (7.0) | 422 (7.9) | 80 (11.3) | 73 (6.9) | 59 (2.9) |
| | **Severe** | 1,609 (17.6) | 1,183 (22.3) | 160 (22.6) | 246 (23.1) | 20 (1.0) |
| **Left Ventricular Hypertrophy, n (%)** | | 2,280 (25.5) | 1,398 (26.9) | 199 (28.6) | 302 (29.0) | 381 (19.1) |
| **Aortic Valve Peak Velocity (m/s), mean (SD)** | | 2.2 (1.2) | 2.4 (1.3) | 2.5 (1.3) | 2.4 (1.3) | 1.6 (0.6) |
| **Interventricular Septum Diastole (cm), mean (SD)** | | 1.1 (0.2) | 1.1 (0.2) | 1.1 (0.2) | 1.1 (0.2) | 1.0 (0.2) |
| **Left Ventricular Posterior Wall Diastole (cm), mean (SD)** | | 1.0 (0.2) | 1.0 (0.2) | 1.1 (0.2) | 1.1 (0.2) | 1.0 (0.2) |
| **Left Ventricular Internal Diameter Diastole (cm), mean (SD)** | | 4.6 (0.7) | 4.6 (0.7) | 4.6 (0.7) | 4.6 (0.7) | 4.6 (0.7) |
| **Left Ventricular Internal Diameter Systole (cm), mean (SD)** | | 3.1 (0.7) | 3.1 (0.7) | 3.1 (0.7) | 3.1 (0.8) | 3.1 (0.7) |
| **Left Ventricular Mass (g), mean (SD)** | | 175.0 (64.2) | 176.5 (63.2) | 178.6 (67.7) | 180.7 (69.4) | 167.0 (61.9) |
| **Left Ventricular Mass Index (g/m^2), mean (SD)** | | 91.0 (29.6) | 92.1 (29.4) | 92.7 (31.0) | 93.8 (32.0) | 85.9 (27.9) |

Descriptive statistics of demographics and label prevalence for each set of the study cohort. Percentages are "valid percentages" calculated for studies with available information. BMI = body mass index; SD = standard deviation; TTE = transthoracic echocardiography.

**Table S2 | Detailed LVH classification results**

| | | AUROC Results on Internal Test Set | | | | |
|---|---|---|---|---|---|---|
| | | **Echo-CLR** | **Kinetics-400** | **Random** | **MI-SimCLR** | **SimCLR** |
| **Train Ratio** | **0.01** | 0.605 (0.572, 0.637) | 0.494 (0.460, 0.528) | 0.462 (0.429, 0.494) | 0.572 (0.540, 0.603) | 0.521 (0.489, 0.552) |
| | **0.05** | 0.669 (0.638, 0.698) | 0.576 (0.545, 0.607) | 0.506 (0.475, 0.538) | 0.632 (0.600, 0.663) | 0.565 (0.532, 0.597) |
| | **0.10** | 0.723 (0.693, 0.751) | 0.605 (0.574, 0.636) | 0.488 (0.456, 0.521) | 0.666 (0.636, 0.696) | 0.619 (0.587, 0.650) |
| | **0.25** | 0.750 (0.723, 0.777) | 0.766 (0.740, 0.792) | 0.625 (0.594, 0.656) | 0.704 (0.674, 0.732) | 0.683 (0.653, 0.713) |
| | **0.50** | 0.769 (0.743, 0.794) | 0.733 (0.705, 0.761) | 0.761 (0.733, 0.787) | 0.746 (0.719, 0.774) | 0.694 (0.664, 0.724) |
| | **1.00** | 0.795 (0.770, 0.819) | 0.807 (0.783, 0.830) | 0.710 (0.683, 0.738) | 0.749 (0.722, 0.776) | 0.740 (0.713, 0.766) |
| | | AUPR Results on Internal Test Set | | | | |
| | | **Echo-CLR** | **Kinetics-400** | **Random** | **MI-SimCLR** | **SimCLR** |
| **Train Ratio** | **0.01** | 0.371 (0.341, 0.407) | 0.311 (0.284, 0.345) | 0.266 (0.250, 0.288) | 0.339 (0.312, 0.372) | 0.299 (0.278, 0.325) |
| | **0.05** | 0.412 (0.382, 0.448) | 0.340 (0.315, 0.371) | 0.292 (0.272, 0.319) | 0.410 (0.376, 0.451) | 0.346 (0.318, 0.380) |
| | **0.10** | 0.489 (0.449, 0.534) | 0.381 (0.349, 0.418) | 0.288 (0.267, 0.315) | 0.427 (0.393, 0.467) | 0.388 (0.356, 0.427) |
| | **0.25** | 0.561 (0.518, 0.605) | 0.554 (0.512, 0.602) | 0.393 (0.361, 0.432) | 0.464 (0.427, 0.506) | 0.458 (0.419, 0.503) |
| | **0.50** | 0.564 (0.522, 0.608) | 0.509 (0.468, 0.556) | 0.554 (0.511, 0.600) | 0.541 (0.501, 0.589) | 0.504 (0.462, 0.547) |
| | **1.00** | 0.615 (0.573, 0.656) | 0.607 (0.564, 0.655) | 0.463 (0.426, 0.505) | 0.542 (0.500, 0.588) | 0.519 (0.477, 0.565) |
| | | AUROC Results on External Test Set | | | | |
| | | **Echo-CLR** | **Kinetics-400** | **Random** | **MI-SimCLR** | **SimCLR** |
| **Train Ratio** | **0.01** | 0.599 (0.573, 0.625) | 0.502 (0.476, 0.531) | 0.467 (0.441, 0.493) | 0.564 (0.537, 0.592) | 0.513 (0.486, 0.540) |
| | **0.05** | 0.677 (0.651, 0.702) | 0.602 (0.575, 0.630) | 0.510 (0.482, 0.537) | 0.640 (0.613, 0.667) | 0.582 (0.555, 0.608) |
| | **0.10** | 0.701 (0.676, 0.725) | 0.605 (0.578, 0.632) | 0.493 (0.465, 0.520) | 0.668 (0.642, 0.695) | 0.627 (0.601, 0.652) |
| | **0.25** | 0.743 (0.720, 0.766) | 0.713 (0.689, 0.737) | 0.625 (0.598, 0.652) | 0.715 (0.691, 0.739) | 0.695 (0.671, 0.720) |
| | **0.50** | 0.782 (0.761, 0.803) | 0.733 (0.710, 0.756) | 0.732 (0.708, 0.756) | 0.741 (0.718, 0.765) | 0.684 (0.659, 0.709) |
| | **1.00** | 0.804 (0.783, 0.824) | 0.806 (0.786, 0.827) | 0.712 (0.688, 0.737) | 0.757 (0.735, 0.779) | 0.742 (0.719, 0.766) |
| | | AUPR Results on External Test Set | | | | |
| | | **Echo-CLR** | **Kinetics-400** | **Random** | **MI-SimCLR** | **SimCLR** |
| **Train Ratio** | **0.01** | 0.266 (0.242, 0.292) | 0.192 (0.180, 0.209) | 0.171 (0.162, 0.183) | 0.234 (0.215, 0.257) | 0.203 (0.188, 0.222) |
| | **0.05** | 0.327 (0.300, 0.360) | 0.274 (0.250, 0.303) | 0.202 (0.186, 0.223) | 0.313 (0.284, 0.347) | 0.243 (0.224, 0.270) |
| | **0.10** | 0.349 (0.318, 0.383) | 0.277 (0.251, 0.308) | 0.192 (0.178, 0.211) | 0.337 (0.306, 0.372) | 0.273 (0.251, 0.301) |
| | **0.25** | 0.415 (0.379, 0.455) | 0.403 (0.368, 0.440) | 0.303 (0.275, 0.337) | 0.373 (0.340, 0.408) | 0.369 (0.335, 0.407) |
| | **0.50** | 0.465 (0.429, 0.505) | 0.436 (0.400, 0.474) | 0.430 (0.393, 0.473) | 0.432 (0.395, 0.470) | 0.367 (0.334, 0.403) |
| | **1.00** | 0.517 (0.478, 0.559) | 0.520 (0.483, 0.561) | 0.412 (0.376, 0.450) | 0.429 (0.393, 0.468) | 0.442 (0.404, 0.483) |

LVH classification results, as measured by AUROC and AUPR, for all fine-tuning ratios on both the internal and external test sets. "Train ratio" refers to the proportion of the available training data used for fine-tuning after initializing the model with the method specific by each column title. Values in parentheses represent 95% confidence intervals determined by bootstrapping the test set. AUPR = are under the precision-recall curve; AUROC = area under the receiver operating characteristic curve; LVH = left ventricular hypertrophy; MI-SimCLR = multi-instance SimCLR.

**Table S3 | Detailed severe AS classification results**

| | | *AUROC Results on Internal Test Set* | | | | |
|---|---|---|---|---|---|---|
| | | **Echo-CLR** | **Kinetics-400** | **Random** | **MI-SimCLR** | **SimCLR** |
| | **0.01** | 0.818 (0.793, 0.840) | 0.612 (0.577, 0.647) | 0.511 (0.477, 0.545) | 0.718 (0.685, 0.751) | 0.569 (0.534, 0.604) |
| | **0.05** | 0.872 (0.853, 0.891) | 0.770 (0.744, 0.796) | 0.638 (0.605, 0.671) | 0.819 (0.794, 0.843) | 0.635 (0.601, 0.669) |
| **Train** | **0.10** | 0.893 (0.876, 0.910) | 0.848 (0.827, 0.869) | 0.851 (0.831, 0.871) | 0.866 (0.846, 0.886) | 0.849 (0.828, 0.870) |
| **Ratio** | **0.25** | 0.904 (0.888, 0.920) | 0.896 (0.878, 0.912) | 0.901 (0.885, 0.916) | 0.903 (0.888, 0.918) | 0.877 (0.858, 0.895) |
| | **0.50** | 0.925 (0.911, 0.938) | 0.928 (0.915, 0.941) | 0.917 (0.903, 0.931) | 0.899 (0.883, 0.915) | 0.879 (0.860, 0.897) |
| | **1.00** | 0.934 (0.920, 0.947) | 0.938 (0.925, 0.951) | 0.925 (0.912, 0.938) | 0.930 (0.917, 0.943) | 0.902 (0.884, 0.918) |
| | | *AUPR Results on Internal Test Set* | | | | |
| | | **Echo-CLR** | **Kinetics-400** | **Random** | **MI-SimCLR** | **SimCLR** |
| | **0.01** | 0.519 (0.475, 0.572) | 0.340 (0.304, 0.385) | 0.230 (0.213, 0.251) | 0.482 (0.434, 0.534) | 0.277 (0.253, 0.312) |
| | **0.05** | 0.628 (0.580, 0.679) | 0.471 (0.428, 0.522) | 0.345 (0.309, 0.387) | 0.533 (0.488, 0.582) | 0.340 (0.305, 0.380) |
| **Train** | **0.10** | 0.666 (0.619, 0.717) | 0.597 (0.548, 0.647) | 0.612 (0.567, 0.660) | 0.603 (0.555, 0.656) | 0.575 (0.528, 0.625) |
| **Ratio** | **0.25** | 0.687 (0.641, 0.740) | 0.691 (0.645, 0.738) | 0.694 (0.650, 0.739) | 0.683 (0.638, 0.729) | 0.666 (0.622, 0.709) |
| | **0.50** | 0.766 (0.725, 0.808) | 0.769 (0.731, 0.809) | 0.735 (0.691, 0.779) | 0.684 (0.638, 0.733) | 0.651 (0.605, 0.697) |
| | **1.00** | 0.818 (0.784, 0.852) | 0.816 (0.780, 0.853) | 0.749 (0.704, 0.795) | 0.769 (0.727, 0.812) | 0.713 (0.670, 0.757) |
| | | *AUROC Results on External Test Set* | | | | |
| | | **Echo-CLR** | **Kinetics-400** | **Random** | **MI-SimCLR** | **SimCLR** |
| | **0.01** | 0.874 (0.820, 0.922) | 0.645 (0.525, 0.761) | 0.437 (0.338, 0.535) | 0.702 (0.610, 0.795) | 0.528 (0.427, 0.628) |
| | **0.05** | 0.935 (0.898, 0.966) | 0.840 (0.760, 0.914) | 0.610 (0.487, 0.730) | 0.920 (0.871, 0.962) | 0.624 (0.507, 0.735) |
| **Train** | **0.10** | 0.943 (0.903, 0.973) | 0.861 (0.788, 0.925) | 0.890 (0.826, 0.943) | 0.946 (0.925, 0.966) | 0.904 (0.863, 0.943) |
| **Ratio** | **0.25** | 0.959 (0.940, 0.977) | 0.960 (0.939, 0.978) | 0.957 (0.928, 0.979) | 0.949 (0.930, 0.967) | 0.962 (0.940, 0.980) |
| | **0.50** | 0.967 (0.943, 0.986) | 0.977 (0.964, 0.988) | 0.969 (0.949, 0.984) | 0.934 (0.901, 0.961) | 0.943 (0.915, 0.968) |
| | **1.00** | 0.947 (0.884, 0.988) | 0.954 (0.921, 0.983) | 0.976 (0.966, 0.984) | 0.964 (0.943, 0.982) | 0.955 (0.933, 0.976) |
| | | *AUPR Results on External Test Set* | | | | |
| | | **Echo-CLR** | **Kinetics-400** | **Random** | **MI-SimCLR** | **SimCLR** |
| | **0.01** | 0.066 (0.040, 0.121) | 0.023 (0.013, 0.048) | 0.008 (0.007, 0.010) | 0.082 (0.018, 0.189) | 0.011 (0.008, 0.016) |
| | **0.05** | 0.194 (0.113, 0.355) | 0.095 (0.049, 0.169) | 0.025 (0.012, 0.062) | 0.245 (0.116, 0.421) | 0.016 (0.011, 0.025) |
| **Train** | **0.10** | 0.190 (0.113, 0.312) | 0.116 (0.042, 0.232) | 0.144 (0.075, 0.286) | 0.243 (0.112, 0.413) | 0.123 (0.062, 0.267) |
| **Ratio** | **0.25** | 0.217 (0.120, 0.356) | 0.165 (0.109, 0.303) | 0.275 (0.153, 0.480) | 0.137 (0.084, 0.238) | 0.294 (0.157, 0.471) |
| | **0.50** | 0.420 (0.254, 0.597) | 0.342 (0.201, 0.519) | 0.270 (0.154, 0.430) | 0.124 (0.074, 0.218) | 0.233 (0.107, 0.383) |
| | **1.00** | 0.337 (0.210, 0.546) | 0.340 (0.187, 0.527) | 0.238 (0.152, 0.399) | 0.296 (0.171, 0.498) | 0.394 (0.228, 0.573) |

Severe AS classification results, as measured by AUROC and AUPR, for all fine-tuning ratios on both the internal and external test sets. "Train ratio" refers to the proportion of the available training data used for fine-tuning after initializing the model with the method specific by each column title. Values in parentheses represent 95% confidence intervals determined by bootstrapping the test set. AS = aortic stenosis; AUPR = are under the precision-recall curve; AUROC = area under the receiver operating characteristic curve; MI-SimCLR = multi-instance SimCLR.

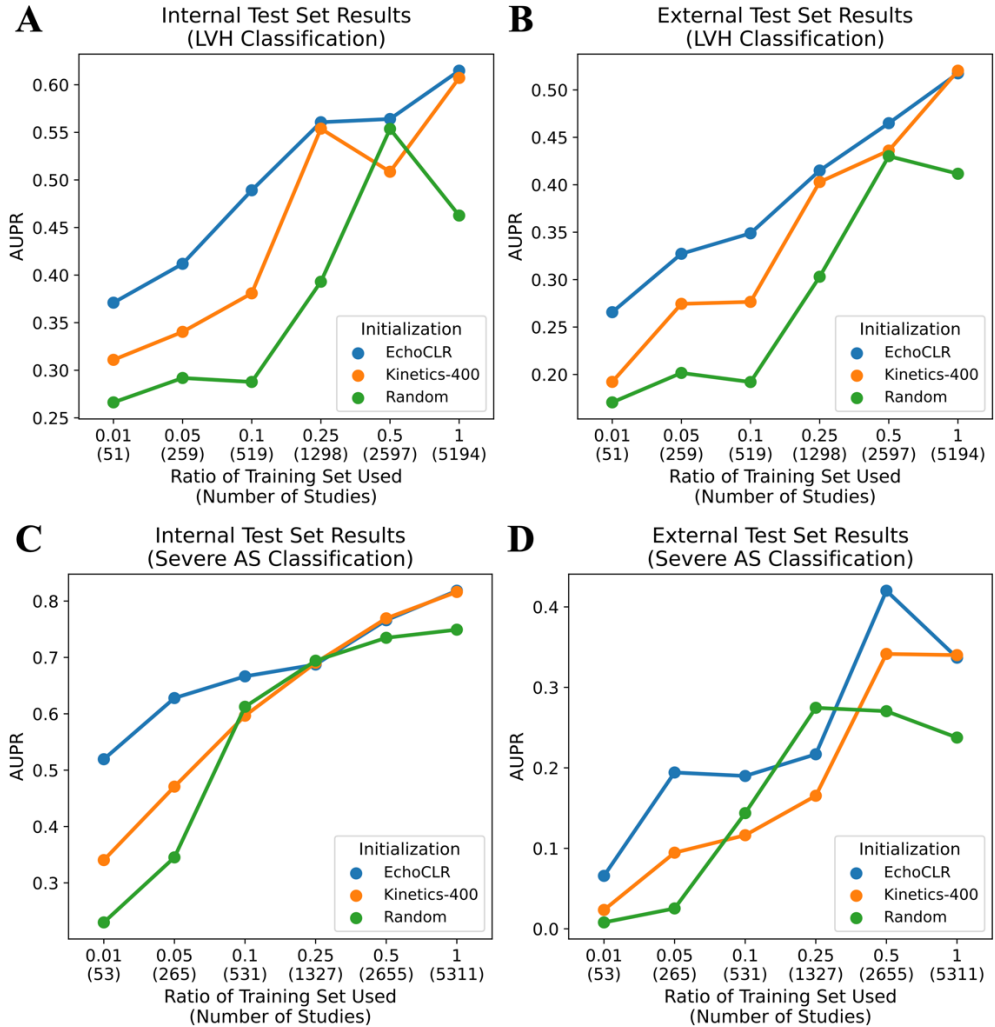

**Fig. S1 | Additional classification performance on different amounts of training data.** AUPR for LVH classification on the internal **(A)** and external test set **(B)** and severe AS classification on the internal **(C)** and external test set **(D)** for a randomly initialized, Kinetics-400-pretrained, and EchoCLR-pretrained model when fine-tuned on different amounts of labeled training data. AS = aortic stenosis; AUPR = area under the precision-recall curve; LVH = left ventricular hypertrophy.

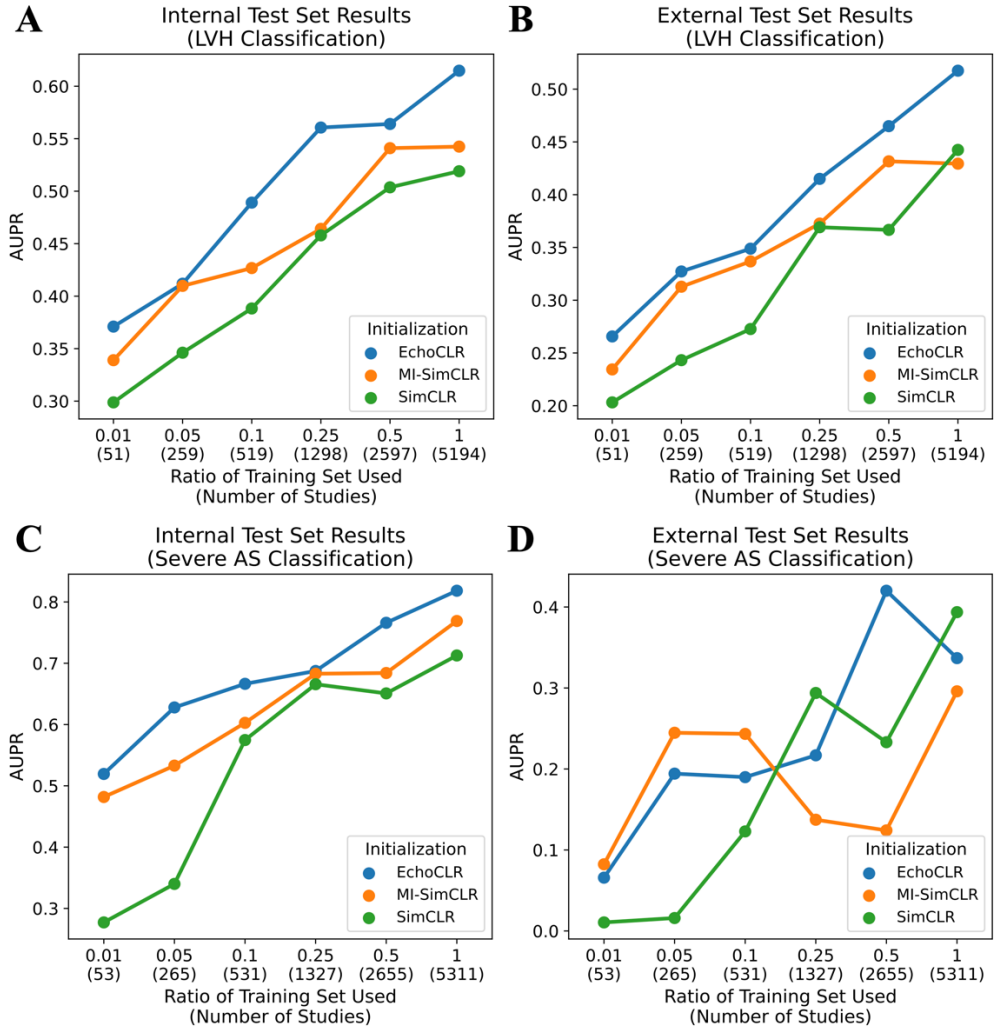

**Fig. S2 | Additional ablation study of EchoCLR when finetuned on different amounts of training data.** AUPR for LVH classification on the internal **(A)** and external test set **(B)** and severe AS classification on the internal **(C)** and external test set **(D)** for a SimCLR-pretrained, MI-SimCLR-pretrained, and EchoCLR-pretrained model when fine-tuned on different amounts of labeled training data. AS = aortic stenosis; AUPR = area under the precision-recall curve; LVH = left ventricular hypertrophy; MI-SimCLR = multi-instance SimCLR.